\documentclass[11pt]{article}
\usepackage[letterpaper,margin=1.02in]{geometry}
\usepackage[T1]{fontenc}
\usepackage{amsmath,amssymb,amsthm,mathtools}
\usepackage{newtxtext,newtxmath}
\usepackage{microtype}
\usepackage{enumitem}
\usepackage{tikz}
\usetikzlibrary{arrows.meta,positioning,calc,decorations.pathreplacing}
\usepackage[round]{natbib}
\usepackage[colorlinks=true,linkcolor=blue!45!black,citecolor=blue!45!black,urlcolor=blue!45!black]{hyperref}
\usepackage{doi}
\usepackage[capitalise,nameinlink,noabbrev]{cleveref}
\hypersetup{
  pdftitle={The Complexity of Oracle-Efficient Realizable Transductive Learning},
  pdfauthor={Changlong Wu},
  pdfkeywords={transductive online learning, realizable prediction, halving, agnostic regret, weighted ERM, oracle complexity}
}

\allowdisplaybreaks[2]
\setlist[itemize]{topsep=3pt,itemsep=2pt,leftmargin=20pt}
\newtheorem{theorem}{Theorem}[section]
\newtheorem{lemma}[theorem]{Lemma}

\theoremstyle{definition}

\theoremstyle{remark}
\newtheorem{remark}[theorem]{Remark}
\newcommand{\cube}{\{0,1\}}
\newcommand{\HH}{\mathcal H}
\newcommand{\EE}{\mathbb E}
\newcommand{\PP}{\mathbb P}
\newcommand{\RR}{\mathbb R}
\newcommand{\one}{\mathbf 1}
\newcommand{\ERM}{\mathsf{ERM}}
\newcommand{\dvc}{d_{\mathrm{VC}}}
\newcommand{\dld}{d_{\mathrm{LD}}}
\DeclareMathOperator*{\argmax}{arg\,max}
\DeclareMathOperator{\lexmin}{lexmin}
\newcommand{\Tn}{\mathrm{T}}
\title{{\Large The Complexity of Oracle-Efficient Realizable Transductive Learning}}

\vspace{1em}
\author{Changlong Wu\\[2pt]
  \normalsize University of Arizona\\[2pt]
  \normalsize\texttt{clwu@arizona.edu}}
\date{}
\begin{document}
\maketitle
\begin{abstract}
In realizable transductive online learning, the learner knows the entire sequence of \(T\) instances in advance, and the labels are consistent with one of \(N\) hypotheses. With explicit access to the hypothesis class, the halving algorithm makes at most \(\log_2 N\) mistakes. \emph{Can a comparable guarantee be achieved using only polynomially many empirical risk minimization (ERM) queries?}
We answer this question negatively, even for a weighted-ERM oracle that optimizes arbitrary real-valued linear objectives and returns a complete maximizing hypothesis. For every fixed \(\beta\in(0,1)\) and every randomized learner making polynomially many oracle queries, there exists a fixed realizable instance with exactly \(N=2^{\lfloor T^\beta\rfloor}\) hypotheses on which the learner makes \(\Omega_\beta(\sqrt{T\log N})\) mistakes in expectation. This matches the agnostic-rate upper bound achieved by random playout using \(O(T)\) oracle calls. The separation persists for much smaller classes as well: along a sequence with \(\log N=\Theta((\log T)^2)\), we establish a mistake lower bound of \(\exp(\Omega(\log T/\log\log T))\).
Our main technical contribution is a \emph{composition lemma} that combines two realizable prediction problems such that their mistake lower bounds add, up to a small loss, while keeping the growth of the combined hypothesis class under control. Iterating this lemma yields a family of hard instances spanning the desired parameter regimes, establishing a sharp separation between information-theoretic and oracle-efficient mistake guarantees.
\end{abstract}

\section{Introduction}\label{sec:intro}

In transductive online learning, the learner knows the entire instance
sequence $x_1,\ldots,x_T$, but not the labels
\citep{ben-david-kushilevitz-mansour,kakade-kalai}.
On each round $t$, the learner predicts a binary label $\hat y_t$ for
$x_t$ and then observes the true label $y_t$.
A \emph{hypothesis class} $\mathcal{H}$ specifies the candidate labelings of these instances.
Since the instance sequence is fixed, we may identify each hypothesis
with its \emph{trace}, the corresponding $T$-bit label vector.
We therefore work directly with a finite class
$\HH\subseteq\cube^T$ of size $N$.
The label sequence is \emph{realizable} if it coincides with some
$h^\star\in\HH$, which we call the \emph{target trace}.

In the realizable setting, the \emph{halving algorithm}
\citep{barzdin1972prediction,littlestone} achieves a mistake upper bound
of $\log_2 N$. It predicts the majority label among hypotheses
consistent with the labels observed so far. Each mistake eliminates
at least half of these hypotheses, yielding the logarithmic upper bound.
The challenge is its computation: implementing the majority prediction
requires counting, or at least comparing the numbers of, consistent
hypotheses supporting each label. This can be difficult when the
hypothesis class is large and represented implicitly.

A natural approach to computational efficiency is to replace explicit
operations over the hypothesis class with calls to an
\emph{optimization oracle} \citep{kalai-vempala,dudik-et-al}.
Empirical risk minimization (ERM) is a particularly natural choice:
it is fundamental to supervised learning and underlies many
oracle-efficient online-learning algorithms
\citep{kakade-kalai,cesabianchi-shamir,wang-hu-muthukumar-abernethy}.
For parameterized models, ERM objectives can often be optimized
approximately using gradient-based methods
\citep{bottou-curtis-nocedal-2018}, while other hypothesis classes
admit specialized optimization procedures.
The ERM-oracle model abstracts away these implementations and measures
the number of optimization queries required for online prediction.
We call a learner \emph{oracle-efficient} if it makes at most
polynomially many oracle queries in the horizon $T$.

Perhaps surprisingly, ERM oracles already suffice for
\emph{agnostic} transductive online learning, where the labels
need not agree with any hypothesis. The random-playout algorithm of
\citet{cesabianchi-shamir-nips11}
(see also \citealp{cesabianchi-shamir,rakhlin-shamir-sridharan})
achieves $O(\sqrt{T\log N})$ expected \emph{regret}, measured
against the best fixed hypothesis in $\HH$, using only two ERM
queries per round. In the realizable setting, the best hypothesis
makes no mistakes, so the same algorithm guarantees
$O(\sqrt{T\log N})$ \emph{expected mistakes}. Thus, polynomially many
ERM queries suffice for sublinear mistakes whenever
$\log N=o(T)$.

This leads to our central question:
\emph{Can realizability improve the mistake bound of an oracle-efficient
transductive learner beyond the agnostic rate?}
In particular, can polynomially many ERM calls recover the logarithmic
bound of halving?  We show that, for general finite hypothesis classes, the aforementioned square-root rate \emph{cannot} be improved in the worst case.

\paragraph{The ERM oracle.}
For any weight vector $w\in\RR^T$, the oracle returns the full
maximizing trace
\begin{equation}\label{eq:erm-intro}
  \ERM_\HH(w):=
  \lexmin\left(\argmax_{h\in\HH}w\cdot h\right),
\end{equation}
where $w\cdot h=\sum_{t=1}^T w_t h_t$ and $\lexmin$ selects the
lexicographically smallest maximizer.
This oracle subsumes weighted ERM for any binary prediction loss
$\ell:\cube\times\cube\to\RR$.
Indeed, given labels $y_1,\ldots,y_T\in\cube$ and nonnegative
weights $a_1,\ldots,a_T$, let
$w_t=a_t\bigl(\ell(0,y_t)-\ell(1,y_t)\bigr)$.
Then
\begin{equation}\label{eq:weighted-erm-reduction}
  \sum_{t=1}^T a_t\ell(h_t,y_t)
  =\sum_{t=1}^T a_t\ell(0,y_t)-w\cdot h,
\end{equation}
so a single oracle call returns a minimizer of the weighted
empirical loss.

Our oracle is deliberately strong: it accepts arbitrary real-valued
weights and reveals the entire optimizing trace. It can therefore
simulate weaker oracle forms that restrict the admissible weights or
reveal only part of the optimizer, provided they use the same
tie-breaking rule. The learner may randomize, issue adaptive queries
before and during prediction, and perform \emph{unlimited computation}
between oracle calls. Its only access to the hypothesis class $\HH$
is through the oracle.

\paragraph{The main result.}
Write
$M_T=\sum_{t=1}^T\mathbf 1\{\widehat y_t\ne h^\star_t\}$
for the learner's number of mistakes.  Throughout, $\log$ means
$\log_2$ and $\ln$ means the natural logarithm.

\begin{theorem}[Polynomial-query lower bound]\label{thm:main}
Fix $\beta\in(0,1)$.  There is a constant $c_\beta>0$ such that for
every fixed $C>0$ and all sufficiently large $T$, with
$N=2^{\lfloor T^\beta\rfloor}$, the following holds.
For every randomized transductive learner $\mathcal A$ using at most
$T^C$ queries to \eqref{eq:erm-intro} on every execution, there are a
fixed class $\HH\subseteq\cube^T$ of exactly $N$ distinct traces and a
fixed $h^\star\in\HH$ for which
\begin{equation}\label{eq:main}
  \EE_\mathcal A M_T\ \ge\ c_\beta\sqrt{T\log N}.
\end{equation}
The pair $(\HH,h^\star)$ may depend on $\mathcal A$; $c_\beta$
does not depend on $C$.
\end{theorem}

The random-playout upper bound discussed above matches \Cref{thm:main} up to
constant factors, establishing that for
$N=2^{\lfloor T^\beta\rfloor}$, the optimal worst-case expected
mistake bound among learners making polynomially many oracle queries
is $\Theta_\beta(\sqrt{T\log N})$, in sharp contrast to the
$O(\log N)$ bound from halving.
The separation persists even for much smaller classes:
\Cref{thm:small} establishes a lower bound of
$\exp(\Omega(\log T/\log\log T))$ mistakes when
$\log N=\Theta((\log T)^2)$, whereas halving makes only
$O((\log T)^2)$ mistakes.

\paragraph{How the proof works.}
We prove the lower bound by constructing a distribution over realizable
pairs $(\HH,h^\star)$; the learner knows the distribution but not its
realization.  The construction combines two independently drawn
pairs $(G,u)$ and $(F,v)$, where
$G\subseteq\cube^s$, $u\in G$, $F\subseteq\cube^r$, and $v\in F$.
The first pair describes the labels in the first $s$ rounds, and the
second describes the labels to be predicted after.  We randomly embed
the $r$ coordinates of $F$ into $D=Br$ new coordinates by an injective
map $E$, and choose a fixed set $C_D\subseteq\cube^D$ of $m$ comparison
traces.  The resulting class and realizable label sequence are
\begin{equation}\label{eq:composition-intro}
  \HH=(G\times C_D)\cup(\{u\}\times E(F)),
  \qquad h^\star=u\Vert E(v),
\end{equation}
where $\Vert$ denotes concatenation.  The first branch consists of all
prefix traces in $G$ paired with traces in $C_D$.  The second
consists of the embedded class $E(F)$, with the \emph{same} realized prefix $u\in G$.  Consequently,
\[
  |\HH|\le |G|\,m+|F|.
\]

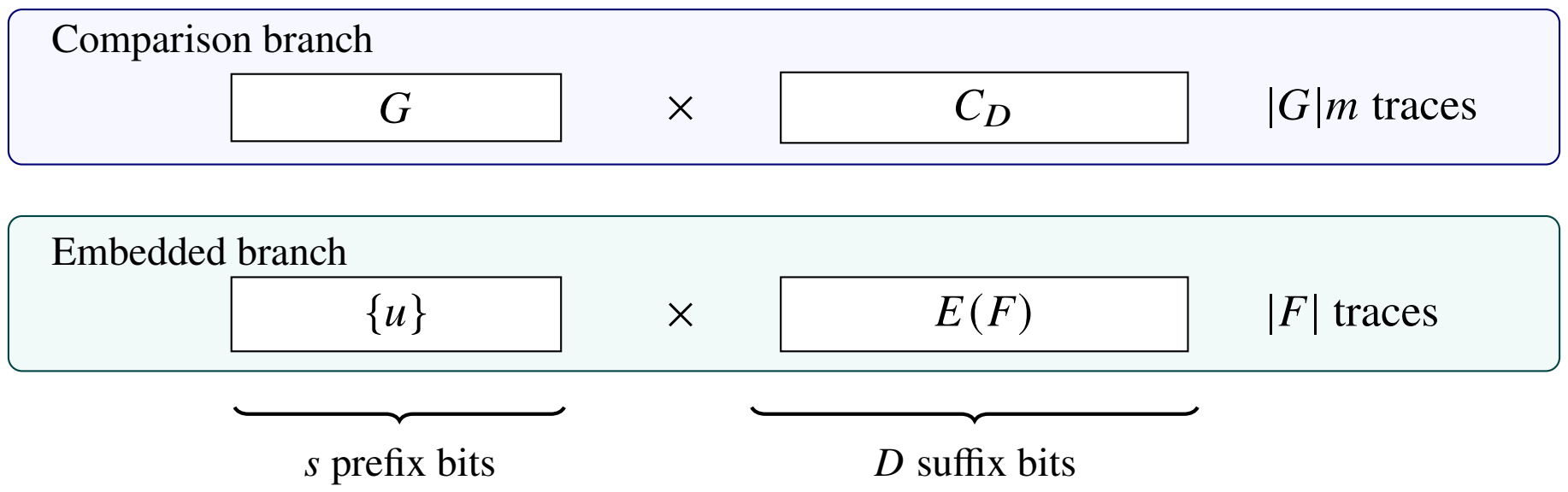
\begin{figure}[t]
\centering
\begin{tikzpicture}[font=\small,>=Latex]
\draw[rounded corners=3pt,fill=blue!3,draw=blue!45!black] (0,1.45) rectangle (12,2.65);
\draw[rounded corners=3pt,fill=teal!5,draw=teal!55!black] (0,-0.15) rectangle (12,1.05);
\node[anchor=west,font=\footnotesize] at (.2,2.39) {Comparison branch};
\node[draw,fill=white,minimum width=2.55cm,minimum height=.52cm] at (3,1.89) {$G$};
\node at (5.2,1.89) {$\times$};
\node[draw,fill=white,minimum width=3.15cm,minimum height=.52cm] at (7.55,1.89) {$C_D$};
\node[anchor=west] at (9.58,1.89) {$|G|m$ traces};
\node[anchor=west,font=\footnotesize] at (.2,.78) {Embedded branch};
\node[draw,fill=white,minimum width=2.55cm,minimum height=.52cm] at (3,.29) {$\{u\}$};
\node at (5.2,.29) {$\times$};
\node[draw,fill=white,minimum width=3.15cm,minimum height=.52cm] at (7.55,.29) {$E(F)$};
\node[anchor=west] at (9.58,.29) {$|F|$ traces};
\draw[decorate,decoration={brace,amplitude=3pt,mirror},thick] (1.75,-.43)--(4.3,-.43) node[midway,below=5pt,font=\footnotesize] {$s$ prefix bits};
\draw[decorate,decoration={brace,amplitude=3pt,mirror},thick] (5.75,-.43)--(9.2,-.43) node[midway,below=5pt,font=\footnotesize] {$D$ suffix bits};
\end{tikzpicture}
\caption{Composition of two realizable class--label pairs.  The
comparison branch contains $G\times C_D$.  The embedded branch
contains the fresh class $E(F)$ only after the realized prefix $u$.
The label sequence $u\Vert E(v)$ belongs to the embedded branch.}
\label{fig:gate}
\end{figure}

Our main goal is to show that the combined problem $(\mathcal{H},h^{\star})$ retains the difficulty of both problems $(G,u)$ and $(F,v)$. We show that $C_D$ can be chosen so that any ERM query returning a trace from $\{u\}\times E(F)$ must induce a halfspace of small uniform measure. A Fourier estimate and Boolean hypercontractivity then show that, for any fixed query, this event occurs with probability at most $\alpha$ over the random embedding $E$ (\Cref{lem:hiding,lem:hitting,lem:leak}). The bound extends to adaptive queries by comparing the actual learner with one whose oracle uses only the first branch. Until their oracle outputs first differ, both learners receive identical information and make identical queries, independently of $E$. Consequently, the first $s$ rounds retain an expected mistake lower bound of $b_L-sQ\alpha$, where $b_L$ is the lower bound for the first prediction problem $(G,u)$ and $Q$ is the number of oracle queries.


For the last $D$ rounds, a learner given the realized prefix $u$ and the embedding $E$ can simulate each ERM query to the combined class using at most one query to $F$. Thus, these rounds retain an expected mistake lower bound of $b_R$, where $b_R$ is the lower bound for the second prediction problem $(F,v)$, without increasing the query budget. Combining the two bounds, the composition lemma yields
\begin{equation}\label{eq:additive-intro}
  \EE M_{s+D}\ge b_L+b_R-sQ\alpha,
  \qquad
  \alpha=\sqrt{2}(D+1)n^{1/(B+1)}m^{-1/2},
\end{equation}
where $n$ is an upper bound on $|F|$. By choosing $m$ sufficiently large, the error $sQ\alpha$ can be made negligible for polynomial $Q$.

Iterating the composition lemma allows mistake lower bounds to accumulate while keeping the logarithm of the class size small. For example, starting with Boolean cubes of dimension $B^2$ under uniformly random labels, a single level of composition combining $B$ copies yields a problem with horizon $T=\Theta(B^4)$, logarithmic class size $\log N=\Theta(B^2)$, and expected mistake lower bound $\Omega(B^3)=\Omega(\sqrt{T\log N})$. Further levels of composition establish the full range of exponents in \Cref{thm:main} and extend the separation to much smaller hypothesis classes in \Cref{thm:small}.

\paragraph{Related work.}
Transductive online learning was studied by
\citet{ben-david-kushilevitz-mansour,kakade-kalai}, and random playout achieves the agnostic-rate upper bound~\citep{cesabianchi-shamir}.
The closest lower bounds are due to
\citet{attias-hanneke-ramaswami}: their ERM lower bound applies to deterministic learners making fewer than $T/2$ queries, while their randomized lower bound allows at most $T/20$ queries to a weaker consistency oracle.
Our result strengthens both bounds, applying to randomized learners making \emph{any fixed polynomial} number of queries, even with access to a weighted-ERM oracle that accepts arbitrary real linear objectives and returns a complete maximizing hypothesis.
In the regime of \Cref{thm:main}, our lower bound matches the agnostic-rate upper bound, establishing that polynomially many ERM queries \emph{cannot} recover the logarithmic mistake bound of halving. This settles the polynomial-query question, while exponentially large query budgets fall outside our scope. Further comparisons appear in \Cref{sec:related}.

\paragraph{Organization.}
We formalize the distributional model in \Cref{sec:model}, prove the
hiding and composition lemmas in
\Cref{sec:hiding,sec:public,sec:composition}, and establish the
main and small-class bounds in
\Cref{sec:recursion,sec:parameters,sec:small}.

\section{Model and distributional lower bounds}\label{sec:model}

\paragraph{Protocol.}
Fix a horizon $T\ge1$.  A finite nonempty class $\HH\subseteq\cube^T$ and
a target $h^\star\in\HH$ are fixed before the interaction begins; the
learner knows $T$ but does not receive $\HH$.  In each round
$t=1,\dots,T$, the learner may issue oracle queries, chosen adaptively on
the basis of its internal randomness, the labels
$h^\star_1,\dots,h^\star_{t-1}$, and all previous oracle answers; it then
predicts $\widehat y_t\in\cube$ and observes $h^\star_t$.  A query
$w\in\RR^T$ is answered with the full trace $\ERM_\HH(w)$ defined in
\eqref{eq:erm-intro}; no other access to $\HH$ is available.  The number of
mistakes is
\[
  M_T=\sum_{t=1}^T\one\{\widehat y_t\ne h^\star_t\}.
\]
Queries made before round~$1$ are permitted and count toward the budget. A learner is a \emph{$Q$-query learner} if it makes at most $Q$ oracle calls on every possible transcript, including those inconsistent with the hypothesis class. This requirement is without loss of generality: any learner satisfying the query bound on realizable transcripts can be truncated after $Q$ calls without changing its behavior on the consistent transcripts. We impose this convention because the proof of \Cref{lem:compose} simulates oracle answers from a class different from the realized one. Learners may be randomized, make predictions outside $\HH$, and perform unlimited computation, including exact real arithmetic, between oracle calls.

\paragraph{Distributional lower bounds.}
We establish lower bounds using distributions $\mathscr D$ over pairs $(\HH,h^\star)$, where $h^\star\in\HH$ almost surely. Define
\begin{equation}\label{eq:LQ}
  L_Q(\mathscr D):=\inf_{\mathcal A}
    \EE_{(\HH,h^\star)\sim\mathscr D,\,\mathcal A}[M_T],
\end{equation}
where the infimum ranges over all $Q$-query learners $\mathcal A$. We allow the learner to know the distribution $\mathscr D$, the horizon and class size, all public parameters of the construction, and the oracle's tie-breaking rule. Only the realized class $\HH$ and target $h^\star$ are hidden. Thus, our lower bounds hold even when the learner knows exactly how the hard instances are generated.

\paragraph{From distributions to fixed instances.}
For technical convenience, our construction first samples a target prefix and then uses it to define part of the hypothesis class. Nevertheless, the entire pair $(\HH,h^\star)$ is sampled before prediction begins; neither the class nor the target changes during learning. Equivalently, one may first sample $\HH$ from its marginal distribution and then $h^\star$ from the corresponding conditional distribution. If $L_Q(\mathscr D)\ge L$, then, by averaging, for every $Q$-query learner $\mathcal A$, there \emph{exists} a fixed pair $(\HH,h^\star)$ in the support of $\mathscr D$ such that
\[
\EE_{\mathcal A}[M_T]\ge L.
\]
Thus, our distributional lower bounds imply worst-case lower bounds for fixed realizable instances. 

\paragraph{The tie-breaking rule.}
We order binary words lexicographically with $0<1$: $x$ precedes $y$ if $x_k<y_k$ at their first differing coordinate. The oracle always returns the lexicographically smallest maximizer, using a fixed rule known to the learner. Specifying this rule is important because the learner can deliberately create ties, for example by querying $w=0$. As shown in \Cref{subsec:ties}, lexicographic tie-breaking is equivalent to maximizing a sufficiently small, class-independent perturbation of the objective. Since a
universal guarantee must hold for every valid implementation of weighted
ERM, our lower bound for this particular implementation rules out such
guarantees. We do not claim that the lower bound holds for arbitrary class-dependent tie-breaking rules, which could reveal additional information through the choice of maximizer. Indeed, \citet{li-2026} shows that tie-breaking can affect the oracle complexity of consistency-type oracles.

\section{Optimizer density under a random embedding}\label{sec:hiding}


\paragraph{The random embedding.}
Fix a nonempty class $F\subseteq\cube^r$ with $|F|\le n$ and an odd integer $B\ge3$, and let $D=Br$. The oddness assumption is needed only for the elementary proof of hypercontractivity in \Cref{app:hyper}. We construct a random embedding $E:F\to\cube^D$ by choosing a uniformly random subset
\[
  S=\{i_1<\cdots<i_r\}\subseteq[D]
\]
of size $r$ and an independent uniform mask $X\in\cube^D$. For every $f\in F$, define
\begin{equation}\label{eq:embedding}
  E(f)=X\oplus\iota_S(f),\qquad
  \iota_S(f)_{i_j}=f_j\ \ (j\in[r]),\qquad
  \iota_S(f)_i=0\ \ (i\notin S),
\end{equation}
where $\oplus$ denotes coordinatewise XOR. The same support $S$ and mask $X$ are used for all $f\in F$. Thus, $E$ is injective, and each fixed $E(f)$ is uniformly distributed over $\cube^D$, although the embedded traces are correlated. For $x\in\cube^D$, we write $x_S=(x_{i_1},\ldots,x_{i_r})$ for its restriction to the coordinates in $S$.

Fix an objective $w\in\RR^D$ independently of $(S,X)$, and let
\[
  Y:=\lexmin\Bigl(\argmax_{y\in E(F)}w\cdot y\Bigr)
\]
be the oracle's output on the embedded class. Although each individual $E(f)$ is uniformly distributed over $\cube^D$, the maximizer $Y$ need not be uniform, since the optimization selects among correlated embedded traces. The following lemma controls this selection bias uniformly over all fixed objectives $w$.

\begin{lemma}[Density bound for the optimizer]\label{lem:hiding}
Let $\mu_D$ be the uniform probability measure on $\cube^D$.  For every set
$A\subseteq\cube^D$ and every fixed $w\in\RR^D$,
\begin{equation}\label{eq:hiding}
  \PP_{S,X}(Y\in A)\ \le\ n^{1/(B+1)}\sqrt{\mu_D(A)}.
\end{equation}
The bound holds for every fixed $F$ with $|F|\le n$.
\end{lemma}


\begin{proof}
All Boolean cubes are equipped with the uniform measure, and we write $\|a\|_p=(\EE|a|^p)^{1/p}$. The proof has four steps.

\emph{Step 1: Bound the density conditional on the support.}
Fix $S$. Since all candidates in $E(F)$ agree with $X$ outside $S$, their weighted scores differ only on $S$. Thus, for fixed $S$, $w$, and $F$, the optimizer $Y_S$ depends only on $X_S$, while $Y_{S^c}=X_{S^c}$ remains independent and uniform. Let
\[
  \varphi_S(z)=2^r\PP(Y_S=z\mid S)
\]
be the density of $Y_S$ relative to the uniform measure on $\cube^r$. Since $Y_S=X_S\oplus f$ for some $f\in F$,
\[
  \{Y_S=z\}\subseteq\bigcup_{f\in F}\{X_S=z\oplus f\}.
\]
Each event in the union has probability $2^{-r}$, so
\begin{equation}\label{eq:dens-bound}
  0\le\varphi_S(z)\le |F|\le n,
  \qquad \EE_z\varphi_S(z)=1.
\end{equation}
Let $\psi$ denote the density of $Y$ relative to the uniform measure $\mu_D$ on $\cube^D$. Conditional on $S$, the coordinates $Y_{S^c}$ are uniformly distributed over $\cube^{S^c}$ and independent of $Y_S$. Thus, the conditional density of $Y$ with respect to $\mu_D$ is $\varphi_S(x_S)$. Averaging over $S$ gives
\begin{equation}\label{eq:g-density}
  \psi(x)=\EE_S\varphi_S(x_S).
\end{equation}
For any $A\subseteq\cube^D$, Cauchy--Schwarz gives
\[
  \PP(Y\in A)
  =\EE_{x\sim\mu_D}[\psi(x)\one_A(x)]
  \le\|\psi\|_2\sqrt{\mu_D(A)}.
\]
It therefore suffices to prove $\|\psi\|_2\le n^{1/(B+1)}$.

\emph{Step 2: Reduce to the intersection of two supports.}
Let $S,S'$ be independent uniform $r$-subsets of $[D]$, and write $I=S\cap S'$. Expanding \eqref{eq:g-density},
\begin{equation}\label{eq:g-square}
  \|\psi\|_2^2
  =\EE_{S,S'}\EE_x[
    \varphi_S(x_S)\varphi_{S'}(x_{S'})
  ].
\end{equation}
Conditional on $x_I$, the two factors $x_S$ and $x_{S'}$ depend on disjoint sets of remaining coordinates and are therefore independent. Writing
\[
  a_I=\EE[\varphi_S(x_S)\mid x_I],
  \qquad
  b_I=\EE[\varphi_{S'}(x_{S'})\mid x_I],
\]
the inner expectation in \eqref{eq:g-square} equals $\EE[a_Ib_I]$. Using $2a_Ib_I\le a_I^2+b_I^2$ and the symmetry between $S$ and $S'$ yields
\begin{equation}\label{eq:g-conditional}
  \|\psi\|_2^2
  \le\EE_{S,S'}
  \bigl\|\EE[\varphi_S(x_S)\mid x_I]\bigr\|_2^2.
\end{equation}

\emph{Step 3: Bound the contribution at each Fourier degree.}
For $U\subseteq[D]$, let
$\chi_U(x)=(-1)^{\sum_{i\in U}x_i}$.
Viewing $\varphi_S(x_S)$ as a function on $\cube^D$, its Fourier expansion is
\[
  \varphi_S(x_S)
  =\sum_{U\subseteq S}\widehat\varphi_S(U)\chi_U(x),\qquad \widehat\varphi_S(U)=\EE_x[\varphi_S(x_S)\chi_U(x)].
\]
Conditioning on $x_I$ keeps only the characters indexed by subsets of
$I$. Hence, Parseval's identity gives
\[
  \bigl\|\EE[\varphi_S(x_S)\mid x_I]\bigr\|_2^2
  =\sum_{U\subseteq I}\widehat\varphi_S(U)^2.
\]
For $U\subseteq S$, the condition $U\subseteq I$ is equivalent to $U\subseteq S'$. Averaging over $S'$ in \eqref{eq:g-conditional}, we obtain
\begin{equation}\label{eq:noise-powers}
  \|\psi\|_2^2
  \le\EE_S\sum_{U\subseteq S}
    \widehat\varphi_S(U)^2\PP(U\subseteq S')
  \le\EE_S\sum_{U\subseteq S}
    B^{-|U|}\widehat\varphi_S(U)^2.
\end{equation}
The last inequality follows because, for $|U|=k\le r$,
\begin{equation}\label{eq:support-inclusion}
  \PP(U\subseteq S')
  =\frac{\binom{D-k}{r-k}}{\binom{D}{r}}
  =\prod_{i=0}^{k-1}\frac{r-i}{D-i}
  \le\left(\frac rD\right)^k
  =B^{-k}.
\end{equation}

\emph{Step 4: Apply hypercontractivity.}
Let $\Tn_\rho$ denote the Boolean noise operator, defined so that $\Tn_\rho\varphi$ has Fourier coefficients
$\widehat{\Tn_\rho\varphi}(U)=\rho^{|U|}\widehat\varphi(U)$.
By Parseval's identity, \eqref{eq:noise-powers} becomes
\[
  \|\psi\|_2^2
  \le\EE_S\|\Tn_{1/\sqrt B}\varphi_S\|_2^2.
\]
The $(p,2)$-hypercontractive inequality
\citep{bonami1970etude}, see also \citet[Chapter~9]{odonnell}, states that
$\|\Tn_\rho\varphi\|_2\le\|\varphi\|_p$
whenever $1\le p\le2$ and $0\le\rho\le\sqrt{p-1}$.
Taking $p=1+1/B$ and $\rho=1/\sqrt B$ gives
\begin{equation}\label{eq:hyper-use}
  \|\Tn_{1/\sqrt B}\varphi\|_2
  \le\|\varphi\|_{1+1/B}.
\end{equation}
A self-contained proof for odd $B$ is given in \Cref{app:hyper}.

Finally, \eqref{eq:dens-bound} implies
\[
  \|\varphi_S\|_p^p
  =\EE\varphi_S^p
  \le n^{p-1}\EE\varphi_S
  =n^{1/B},
\]
and therefore
\[
  \|\varphi_S\|_p\le n^{1/(B+1)}.
\]
Combining this with \eqref{eq:noise-powers} and \eqref{eq:hyper-use} yields
$\|\psi\|_2^2\le n^{2/(B+1)}$.
The bound \eqref{eq:hiding} now follows from Step~1.
\end{proof}

\section{Comparison sets for linear queries}\label{sec:public}
Let $C_D\subseteq\cube^D$ be a fixed set of comparison traces. For a linear query, an optimizer over the embedded class $E(F)$ can outperform every element of $C_D$ only if it lies in a halfspace disjoint from $C_D$. We therefore construct $C_D$ to intersect every halfspace of sufficiently large uniform measure. The existence of such a set follows from a probabilistic argument using the Sauer--Shelah lemma \citep{sauer1972density}, since affine halfspaces in $\RR^D$ have VC dimension at most $D+1$.

\begin{lemma}[A public comparison set]\label{lem:hitting}
For every integer $1\le m\le2^D$ there is a set $C_D\subseteq\cube^D$ of
exactly $m$ traces that intersects every strict affine halfspace
$A=\{x\in\cube^D:w\cdot x>a\}$, with $w\in\RR^D$ and $a\in\RR$, for which
\begin{equation}\label{eq:threshold}
  \mu_D(A)>\varepsilon_D,\qquad \varepsilon_D:=\frac{2(D+1)^2}{m}.
\end{equation}
\end{lemma}

\begin{proof}
If $\varepsilon_D\ge1$, the claim is immediate, since every subset of $\cube^D$ has uniform measure at most $1$. Otherwise, sample $m$ distinct points uniformly without replacement from $\cube^D$. A fixed set $A\subseteq\cube^D$ is missed with probability
\[
  \binom{2^D-|A|}{m}\Big/\binom{2^D}{m}
  \le (1-\mu_D(A))^m
  \le e^{-m\mu_D(A)}.
\]

To apply a union bound over all halfspaces, we first bound the number of distinct subsets they induce on $\cube^D$. Since strict affine halfspaces in $\RR^D$ have VC dimension at most $D+1$, the Sauer--Shelah lemma \citep{sauer1972density} implies that the number of distinct intersections with $\cube^D$ is at most
\[
  \sum_{k=0}^{D+1}\binom{2^D}{k}
  \le (2^D+1)^{D+1}.
\]
Consequently, by a union bound, the probability that the sampled set misses some halfspace of uniform measure greater than $\varepsilon_D$ is at most
\[
  (2^D+1)^{D+1}e^{-m\varepsilon_D}
  =(2^D+1)^{D+1}e^{-2(D+1)^2}
  <1,
\]
where the last inequality follows from $\ln(2^D+1)<2(D+1)$. Therefore, there exists a fixed set $C_D$ of size $m$ intersecting every halfspace of uniform measure greater than $\varepsilon_D$.
\end{proof}

\begin{lemma}[Probability of outperforming the comparison set]\label{lem:leak}
Let $E$ be the random embedding of \Cref{sec:hiding}, with $|F|\le n$, and
let $C_D$ be as in \Cref{lem:hitting}.  For every fixed $w\in\RR^D$ chosen
independently of the embedding,
\begin{equation}\label{eq:leak}
  \PP\Bigl(\max_{f\in F}w\cdot E(f)>\max_{c\in C_D}w\cdot c\Bigr)
  \ \le\ \alpha(n,D,m)
  :=\sqrt2\,(D+1)\,n^{1/(B+1)}\,m^{-1/2}.
\end{equation}
Here $\alpha$ also depends on the expansion factor $B=D/r$, which we
suppress from the notation.
\end{lemma}

\begin{proof}
Let $b_w=\max_{c\in C_D}w\cdot c$ and $A_w=\{x\in\cube^D:w\cdot x>b_w\}$.
Since $A_w$ is disjoint from $C_D$, \Cref{lem:hitting} gives
$\mu_D(A_w)\le\varepsilon_D$. Let $Y$ be the optimizer over $E(F)$. The event in \eqref{eq:leak} is precisely $\{Y\in A_w\}$. 
By \Cref{lem:hiding} and \eqref{eq:threshold}, its probability is at most
$n^{1/(B+1)}\sqrt{\varepsilon_D}=\alpha(n,D,m)$.
\end{proof}

Writing $g=\log n$ and $a=\log m$, the bound is
\begin{equation}\label{eq:alpha-exp}
  \alpha(n,D,m)=\sqrt2\,(D+1)\,2^{g/(B+1)-a/2}.
\end{equation}
In particular, $a\ge4g/B$ implies
$\alpha(n,D,m)\le\sqrt2(D+1)2^{-g/B}$.
The parameter choices in \Cref{sec:parameters} make this smaller than every fixed
inverse polynomial in the final horizon.

\section{Adaptive composition}\label{sec:composition}

\Cref{lem:leak} controls a \emph{fixed} query.  On its own it does not
handle an adaptive learner, whose later queries may depend on earlier oracle answers. To handle adaptivity, we compare the actual execution with a reference execution whose oracle is restricted to the first branch. We also show that predicting the last $D$ labels remains at least as hard as the original prediction problem on $F$. Together, these two arguments establish that the mistake lower bounds of both components are preserved under composition.

\subsection{Lexicographic tie-breaking}\label{subsec:ties}

Tie-breaking creates two difficulties in the reductions. First, the full oracle and the comparison-only oracle may return different maximizers with equal scores, so their disagreement need not satisfy the strict inequality in \Cref{lem:leak}. Second, XOR masking does not preserve lexicographic order, so the inner oracle may break ties differently from the outer oracle. We resolve both issues by implementing lexicographic tie-breaking through a sufficiently small, class-independent perturbation of the objective in the full outer dimension.

\begin{lemma}[Class-independent lexicographic perturbation]\label{lem:ties}
For every dimension $L$ and every $w\in\RR^L$ there is $\delta>0$,
depending only on $(L,w)$, such that the refined objective
\begin{equation}\label{eq:refine}
  \widetilde w_i=w_i-\delta 2^{-i},\qquad i\in[L],
\end{equation}
assigns distinct scores to distinct words of $\cube^L$.  In particular, for
every nonempty $K\subseteq\cube^L$, the objective $\widetilde w$ has a unique
maximizer over $K$, and this maximizer is $\ERM_K(w)$.
\end{lemma}

\begin{proof}
Let $\kappa(x)=\sum_{i=1}^L2^{-i}x_i$.  If $x\ne y$ first differ at
coordinate $k$, then $\kappa(x)-\kappa(y)$ has the sign of $x_k-y_k$,
because $2^{-k}>\sum_{i=k+1}^L2^{-i}$.  Thus $\kappa$ strictly represents
the lexicographic order, and $|\kappa(x)-\kappa(y)|<1$ for all $x,y$.  If
not all words have the same $w$-score, let $\gamma>0$ be the minimum of
$|w\cdot x-w\cdot y|$ over pairs with different scores (it exists because
the cube is finite) and set $\delta:=\gamma/2$; otherwise set $\delta:=1$.
Since $\widetilde w\cdot x=w\cdot x-\delta\kappa(x)$, the perturbation
preserves every strict comparison of $w$-scores and breaks every tie in
favor of the lexicographically smaller word.  Hence distinct words have
distinct refined scores, and the unique $\widetilde w$-maximizer over $K$
is the lexicographically smallest $w$-maximizer.
\end{proof}

The refinement is an analytical device: the learner still asks $w$, but a
simulator may compute the unique maximizer under $\widetilde w$, which is
exactly the oracle's answer to $w$.  Since $\delta$ is determined by the
whole cube rather than by the unknown class, computing it requires no
information about the class.  In the suffix reduction below it is
essential to refine first, in the full dimension, and only then restrict
to the masked support.

\subsection{The additive composition lemma}\label{subsec:compose-lemma}

Let $\mathscr D_L$ and $\mathscr D_R$ be distributions over pairs $(G,u)$
and $(F,v)$ with $G\subseteq\cube^s$ and $F\subseteq\cube^r$, and suppose
that $|F|\le n$ almost surely.  Fix an odd $B\ge3$, set $D=Br$, and fix a
public $m$-trace comparison set $C_D$ given by \Cref{lem:hitting}; $B$ and
$m$ are fixed throughout this section.  Draw $(G,u)\sim\mathscr D_L$,
$(F,v)\sim\mathscr D_R$ and the embedding $(S,X)$ of \eqref{eq:embedding}
mutually independently.  The \emph{composed distribution}
$\mathscr D_L\star\mathscr D_R$ is the law of
\begin{equation}\label{eq:composition}
  \HH=(G\times C_D)\cup\bigl(\{u\}\times E(F)\bigr),
  \qquad h^\star=u\Vert E(v).
\end{equation}
It has horizon $s+D$, it is always realizable, and $|\HH|\le|G|m+n$.

\begin{lemma}[Additive composition]\label{lem:compose}
Let $Q\ge1$.  If $L_Q(\mathscr D_L)\ge b_L$ and $L_Q(\mathscr D_R)\ge b_R$,
then
\begin{equation}\label{eq:compose-bound}
  L_Q(\mathscr D_L\star\mathscr D_R)\ \ge\ b_L+b_R-sQ\,\alpha(n,D,m).
\end{equation}
\end{lemma}


\begin{proof}
Fix a $Q$-query learner $\mathcal A$ for the composed prediction problem, and write
$M_{\mathrm{prefix}}$ and $M_{\mathrm{suffix}}$ for its numbers of mistakes
in rounds $1,\dots,s$ (the \emph{prefix}) and $s+1,\dots,s+D$ (the
\emph{suffix}).  We bound their expectations separately.

\smallskip\noindent\emph{Part I: a reference execution.}
Consider the \emph{reference execution} of $\mathcal A$, in which the prefix
labels are the true labels $u_1,\dots,u_s$ but every oracle query is
answered from the comparison branch $G\times C_D$ instead of the full class
$\HH$.  A learner $\mathcal A_L$ for the problem $\mathscr{D}_L$ can simulate the prefix of
this execution with one query to $\ERM_G$ per outer query.  Indeed, for
$w=(w_L,w_R)$ with $w_L\in\RR^s$ and $w_R\in\RR^D$,
\begin{equation}\label{eq:public-separate}
  \max_{(x,c)\in G\times C_D}w\cdot(x,c)
  =\max_{x\in G}w_L\cdot x+\max_{c\in C_D}w_R\cdot c,
\end{equation}
and the lexicographically smallest maximizer over the product is the
concatenation of the lexicographically smallest maximizers over the two
factors.  Since $C_D$ is public, the second factor requires no oracle
calls, and $\mathcal A_L$ uses each left label only after it is revealed.
Because the query bound holds on every transcript (\Cref{sec:model}), $\mathcal A_L$ makes
at most $Q$ queries even though the comparison-only query outputs need not come
from the realized class.  Thus $\mathcal A_L$ is a $Q$-query learner for
$\mathscr D_L$ whose mistakes are exactly the reference execution's prefix
mistakes, and
\begin{equation}\label{eq:reference-mistakes}
  \EE\bigl[M^{\mathrm{ref}}_{\mathrm{prefix}}\bigr]\ \ge\ b_L.
\end{equation}

\smallskip\noindent\emph{Part II: coupling the adaptive executions.}
Call a query a \emph{prefix query} if it is issued before the prediction
$\widehat y_s$; only prefix queries can influence prefix predictions.  Run
the reference and actual executions with the same random tape; both see
the same prefix labels $u$.  Condition on $(G,u)$, on the tape, and on
$(F,v)$.  Then the reference execution's prefix queries
$w^{(1)},\dots,w^{(K)}$, $K\le Q$, are fixed---including queries chosen
adaptively after earlier prefix labels---while $(S,X)$ remains uniformly
distributed and independent of them.  Let $\mathcal L_i$ be the event that
the full oracle's output to $w^{(i)}$ differs from the comparison-only output.

Fix $i$, and let $\widetilde w$ be the refinement of $w=w^{(i)}$ from
\Cref{lem:ties} in dimension $s+D$.  On $\mathcal L_i$, the unique
$\widetilde w$-maximizer over $\HH$ does not belong to the comparison branch;
hence it equals $(u,E(f))$ for some $f\in F$, and it strictly beats every
comparison-branch trace under $\widetilde w$.  Since $(u,c)$ is a comparison-branch trace for every
$c\in C_D$,
\begin{align}
  \widetilde w\cdot(u,E(f))
    &>\widetilde w\cdot(u,c)\qquad\text{for all }c\in C_D,\notag\\
  \text{and hence}\qquad
  \widetilde w_R\cdot E(f)
    &>\max_{c\in C_D}\widetilde w_R\cdot c.\label{eq:strict-improvement}
\end{align}
The suffix objective $\widetilde w_R$ is determined by $w^{(i)}$, not by the
embedding, so \Cref{lem:leak} applied to $\widetilde w_R$ gives
$$\PP(\mathcal L_i\mid G,u,\text{tape},F,v)\le\alpha(n,D,m).$$

If the actual and reference prefix transcripts differ, consider the first
prefix query at which their outputs differ.  All earlier outputs and
labels agree, so this query is $w^{(i)}$ for some $i$, and $\mathcal L_i$
occurs.  (We never apply the fixed-query bound conditionally on a history that
contains query outputs from the embedded branch.)  By a union bound over
$i\le K\le Q$,
\begin{equation}\label{eq:disagree}
  \PP(\text{some prefix query output differs})\ \le\ Q\,\alpha(n,D,m).
\end{equation}
Without a disagreement, the two executions make identical prefix
predictions; with one, their prefix mistake counts differ by at most $s$.
By \eqref{eq:reference-mistakes} and \eqref{eq:disagree},
\begin{equation}\label{eq:prefix-lb}
  \EE\bigl[M^{\mathrm{act}}_{\mathrm{prefix}}\bigr]\ \ge\ b_L-sQ\,\alpha(n,D,m).
\end{equation}

\smallskip\noindent\emph{Part III: simulate the composed learner using the right oracle.}
For the suffix, we use a separate reduction: a learner $\mathcal A_R$
for the problem $\mathscr D_R$ that simulates the \emph{actual} composed execution.
The learner $\mathcal A_R$ samples $(G,u)\sim\mathscr D_L$ and $(S,X)$
itself, so it knows them completely, and it knows the public set $C_D$; it
accesses $F$ only through $\ERM_F$ and the sequentially revealed labels of
$v$.

\emph{Oracle outputs.}  Given an outer query $w=(w_L,w_R)$, the learner
$\mathcal A_R$ refines the full objective to $\widetilde w$ (\Cref{lem:ties})
and computes the maximizer over $G\times C_D$ locally.
For a hypothesis in the embedded branch, the bit identity
$b\oplus x=x+(1-2x)b$ yields
\begin{equation}\label{eq:mask-identity}
  \widetilde w\cdot(u,E(f))
  =\widetilde w_L\cdot u+\widetilde w_R\cdot X
   +\sum_{j=1}^{r}(1-2X_{i_j})\,\widetilde w_{R,i_j}\,f_j,
\end{equation}
where $\widetilde w_{R,i}$ denotes the coefficient of suffix coordinate $i$
in the refined $(s+D)$-dimensional objective.  The learner $\mathcal A_R$
issues the single query
\begin{equation}\label{eq:inner-query}
  \lambda_j=(1-2X_{i_j})\,\widetilde w_{R,i_j},\qquad j\in[r],
\end{equation}
to $\ERM_F$.  By \eqref{eq:mask-identity}, $\lambda\cdot f$ differs from the
refined score of $(u,E(f))$ by a constant that does not depend on $f$.
Since $E$ is injective and distinct words have distinct refined scores
(\Cref{lem:ties}), the map $f\mapsto\lambda\cdot f$ is injective on $F$, so
the maximizer of $\lambda\cdot f$ over $F$ is unique and the inner oracle's
tie-breaking rule is never invoked---even though the XOR mask may reverse
lexicographic order.  Finally, $\mathcal A_R$ returns whichever of the best
trace in the comparison branch and the best trace in the embedded
branch has the larger refined score.  Equal scores can occur only
when the two traces coincide.  This reproduces the
outer oracle's answer exactly, using exactly one inner call.

\emph{Labels and predictions.}  The learner $\mathcal A_R$ knows every prefix
label (the prefix target is $u$) and every suffix label off the support
(the suffix label at $i\notin S$ is $X_i$).  At support coordinate $i_j$ the
suffix label is $X_{i_j}\oplus v_j$.  When the simulated learner predicts
$\widehat y_{s+i_j}$, the learner $\mathcal A_R$ predicts
$\widehat v_j=\widehat y_{s+i_j}\oplus X_{i_j}$, observes $v_j$, and feeds
back $v_j\oplus X_{i_j}$ as the outer label.  The mistake indicators
coincide:
\begin{equation}\label{eq:mistake-indicator}
  \one\{\widehat v_j\ne v_j\}=\one\{\widehat y_{s+i_j}\ne v_j\oplus X_{i_j}\}.
\end{equation}
Because $i_1<\cdots<i_r$, the learner $\mathcal A_R$ never needs $v_j$ before
making its own $j$th prediction.  All of its oracle calls, including
those that simulate outer queries during the prefix, are made to
$\ERM_F$, so $\mathcal A_R$ is a $Q$-query learner for $\mathscr D_R$.  Its
simulated execution has exactly the law of the actual composed
execution, and its mistakes are the actual execution's mistakes at the
$r$ support coordinates of the suffix.  Hence
\begin{equation}\label{eq:suffix-lb}
  \EE\bigl[M^{\mathrm{act}}_{\mathrm{suffix}}\bigr]\ \ge\ b_R.
\end{equation}
Adding \eqref{eq:prefix-lb} and \eqref{eq:suffix-lb} proves
\eqref{eq:compose-bound}.
\end{proof}

\begin{remark}[The role of the tie-breaking rule]
The proof uses the same known lexicographic rule for every version of the
oracle.  The refinement of \Cref{lem:ties} turns a changed answer into a
\emph{strict} improvement and, independently, makes the masked-suffix
simulation exact.  Assuming that ties occur with probability zero would be
incorrect: the learner can query the zero vector or create ties
deliberately.
\end{remark}

\section{Recursive amplification: mistakes versus class size}\label{sec:recursion}

We now iteratively apply the composition lemma to amplify the mistake lower bound while controlling the growth of the hypothesis class. We refer to each application as an \emph{append}, since it places a new prediction problem after the existing one.

\subsection{The recursive distributions}

Fix a base dimension $d\ge1$, a number of copies $q\ge2$ per level, a depth
$J\ge1$, and an odd expansion factor $B\ge3$, all integers.  The base
distribution $\mathscr D_0$ has the full cube $\cube^d$ as its class and a
uniformly random target in that cube.  Its horizon is $t_0=d$.  Since the hypothesis class is fixed, oracle answers provide no additional information about the target beyond the previously observed labels, even when queries are chosen adaptively. Each unrevealed target bit therefore remains an independent fair coin, so every learner makes exactly $d/2$ mistakes in expectation.

At level $j\in[J]$, suppose $\mathscr D_{j-1}$ has horizon $t_{j-1}$.
Draw $q$ independent instances from this distribution.  Start with the first and append the
remaining $q-1$ one at a time using \eqref{eq:composition}.  Each append
uses a fresh embedding into $Bt_{j-1}$ suffix coordinates and a fixed public
comparison set of size $m_j=2^{a_j}$, where $a_j$ is an integer satisfying
$1\le a_j\le Bt_{j-1}$.  Write $\star_j$ for this composition of
distributions, with independent factors.  If $\mathscr P_{j,k}$ denotes the
law after combining the first $k$ instances, then
\[
  \mathscr P_{j,1}=\mathscr D_{j-1},\qquad
  \mathscr P_{j,k+1}=\mathscr P_{j,k}\star_j\mathscr D_{j-1}
    \quad(1\le k<q),\qquad
  \mathscr D_j=\mathscr P_{j,q}.
\]
The same comparison set is used for all appends at level $j$.

Overlaps between the comparison and embedded branches can reduce the
number of distinct traces.  Write $n_j$ for the \emph{nominal class size}
at level $j$, obtained by counting the two branches separately at every
append.  Every class drawn from $\mathscr D_j$ has at most $n_j$ traces,
with equality if there are no overlaps in its construction.

\subsection{Three recurrences}

\paragraph{Horizon.}
The first copy contributes $t_{j-1}$ coordinates, and each of the other
$q-1$ copies is expanded to $Bt_{j-1}$ coordinates.  Thus
\begin{equation}\label{eq:horizon}
  t_0=d,\qquad t_j=\bigl[1+(q-1)B\bigr]\,t_{j-1},\qquad
  t_J=d\,\bigl[1+(q-1)B\bigr]^J.
\end{equation}
Write $\tau:=t_J$ for the horizon before we pad the construction to a
prescribed $T$ in \Cref{sec:parameters}.

\paragraph{Class size.}
Suppose the accumulated class has nominal size $P$. Each append creates a comparison branch of nominal size $Pm_j$ and an embedded branch of nominal size $n_{j-1}$. Thus, the nominal class size evolves according to $P\mapsto Pm_j+n_{j-1}$.
Starting from $P=n_{j-1}$ and applying this recurrence $q-1$ times gives
\begin{equation}\label{eq:class-recur}
  n_0=2^d,\qquad n_j=n_{j-1}\bigl(1+m_j+\cdots+m_j^{q-1}\bigr).
\end{equation}
Let $g_j=\log n_j$.  Since $m_j=2^{a_j}\ge2$, we have
$m_j^{q-1}\le\sum_{k=0}^{q-1}m_j^k\le2m_j^{q-1}$, and hence
\begin{equation}\label{eq:logsize}
  g_{j-1}+(q-1)a_j\ \le\ g_j\ \le\ g_{j-1}+(q-1)a_j+1.
\end{equation}
Thus, each level increases the logarithmic class size by approximately $(q-1)a_j$.

\paragraph{Mistakes.}
Each base instance contributes $d/2$ to the lower bound.  Each append
subtracts the error term of \Cref{lem:compose}.  The right factor at
level $j$ is $\mathscr D_{j-1}$, with class size at most $n_{j-1}$, so the
probability bound for a fixed query is
\begin{equation}\label{eq:alpha-j}
  \alpha_j:=\alpha(n_{j-1},Bt_{j-1},m_j)
  =\sqrt2\,(Bt_{j-1}+1)\,2^{g_{j-1}/(B+1)-a_j/2}.
\end{equation}
The left factor $\mathscr P_{j,k}$ has horizon $t_{j-1}(1+(k-1)B)<t_j$.
Applying \Cref{lem:compose} to each of the $q-1$ appends with the same
query budget $Q$ gives
\[
  L_Q(\mathscr D_j)\ \ge\ q\,L_Q(\mathscr D_{j-1})-(q-1)\,t_jQ\,\alpha_j.
\]
Since $L_Q(\mathscr D_0)=d/2$, induction on $j$ gives
\[
  L_Q(\mathscr D_J)\ \ge\ \frac{dq^J}{2}-Q\sum_{j=1}^J(q-1)q^{J-j}t_j\alpha_j
  \ \ge\ \frac{dq^J}{2}-(q^J-1)\,\tau Q\max_{1\le j\le J}\alpha_j.
\]
Since $q^J\le\tau$ by \eqref{eq:horizon}, we obtain
\begin{equation}\label{eq:master}
  L_Q(\mathscr D_J)\ \ge\ \frac{dq^J}{2}-\tau^2Q\max_{1\le j\le J}\alpha_j.
\end{equation}
Increasing $m_j$ improves the fixed-query bound in \eqref{eq:alpha-j}
and raises the class-size cost in \eqref{eq:logsize}.  We balance these
two effects in the parameter choices below.

\subsection{Making all class sizes exact}\label{subsec:collisions}

The nominal class size may overcount hypotheses if the two branches overlap. We show that such overlaps are sufficiently unlikely to be eliminated by conditioning, without significantly weakening the mistake lower bound.
Consider an append at level $j$, with right class $F$ and suffix dimension $D=Bt_{j-1}$. Since the two branches share only the realized prefix $u$, an overlap occurs precisely when $C_D\cap E(F)\ne\varnothing$. Conditional on $F$, each $E(f)$ is uniform on $\cube^D$. A union bound therefore gives
\begin{equation}\label{eq:collision-one}
  \PP\bigl(C_D\cap E(F)\ne\varnothing\bigm| F\bigr)
  \le m_j|F|\,2^{-D}
  \le 2^{a_j+g_{j-1}-Bt_{j-1}}.
\end{equation}

The full recursive construction contains
$(q-1)\sum_{j=1}^J q^{J-j}=q^J-1$ appends.
Let $\mathcal E$ denote the event that no overlap occurs at any append. By another union bound,
\begin{equation}\label{eq:collision-all}
  \zeta:=\PP(\mathcal E^c)
  \le(q^J-1)\max_{j\in[J]}
  2^{a_j+g_{j-1}-Bt_{j-1}}.
\end{equation}
On $\mathcal E$, the resulting class has exactly $n_J$ distinct traces.

We now show that conditioning on $\mathcal E$ preserves the lower bound up to a negligible error. Let $\tau=t_J$ denote the horizon of the final composed problem, and let $M_\tau$ be the total number of mistakes over its $\tau$ rounds. Fix any $Q$-query learner for the conditioned distribution, and run the same learner on the unconditioned distribution, with the same announced horizon and class size. Since the query cap holds on every transcript by \Cref{sec:model}, the lower bound \eqref{eq:master} applies to this learner even on instances where overlaps occur. Moreover, since $0\le M_\tau\le\tau$, for $\zeta<1$,
\begin{equation}\label{eq:conditioning}
  \EE[M_\tau\mid\mathcal E]
  \ge\frac{\EE[M_\tau]-\tau\zeta}{1-\zeta}
  \ge\EE[M_\tau]-\frac{\tau\zeta}{1-\zeta}.
\end{equation}
The parameter choices below ensure $\zeta=o(1/\tau)$. Thus, conditioning on $\mathcal E$ guarantees exactly $n_J$ hypotheses while reducing the mistake lower bound by only $o(1)$. Importantly, this conditioning is performed only after establishing the lower bound for the unconditioned recursive distribution.

\subsection{A worked one-level example}\label{subsec:example}

A single level already illustrates the result.
Let $B\to\infty$ through odd integers, and take
$J=1$, $d=B^2$, $q=B$ and $a_1=4B$.  We combine $B$ full cubes of
dimension $B^2$ with independent random targets.  The horizon is
\[
  \tau=B^2\bigl[1+(B-1)B\bigr]=\Theta(B^4).
\]
The nominal class size is
\[
  n_1=2^{B^2}\bigl(1+2^{4B}+\cdots+2^{4B(B-1)}\bigr),\qquad
  \log n_1=B^2+4B(B-1)+O(1)=\Theta(B^2).
\]
The fixed-query bound is $\alpha_1\le\sqrt2\,(B^3+1)2^{-B}$, and the
overlap probability is at most $(B-1)2^{4B+B^2-B^3}$.  Combining
\eqref{eq:master} and \eqref{eq:conditioning}, every learner making
polynomially many queries incurs at least $B^3/2-o(1)$ expected mistakes
under the distribution conditioned on no overlaps.  Thus
$\log n_1=\Theta(\tau^{1/2})$ and the lower bound is
$\Omega(\tau^{3/4})=\Omega(\sqrt{\tau\log n_1})$. This illustrates the scaling of \Cref{thm:main} for $\beta=\frac12$; the exact horizon and class-size requirements are handled in \Cref{sec:parameters}.


\section{Matching the agnostic square-root rate}\label{sec:parameters}

We now choose the recursion parameters to match the agnostic square-root rate. 
At each level, we set $q=B$. Meanwhile, we choose the comparison set sizes so that the logarithm of the class size remains $\Theta(d)$ across all levels. Starting from $d/2$ expected mistakes over $d$ rounds, these scalings yield a mistake lower bound of order $\sqrt{\tau d}=\Theta(\sqrt{\tau\log n_J})$.

\subsection{Choosing the depth and comparison-set sizes}

Fix $\beta\in(0,1)$, and set
\begin{equation}\label{eq:Jchoice}
  J:=\Bigl\lfloor\frac{1-\beta}{2\beta}\Bigr\rfloor+1,\qquad
  \theta:=\frac{1-\beta}{2\beta J}\in(0,1).
\end{equation}
The argument works for any integer $J>(1-\beta)/(2\beta)$.  As $d\to\infty$,
let $B$ be the least odd integer at least $d^\theta$, so $B\sim d^\theta$,
and set $q=B$.  Choose the comparison-set exponent
\begin{equation}\label{eq:achoice}
  a_j=\Bigl\lceil\frac{4g_{j-1}}B\Bigr\rceil.
\end{equation}
Here $g_{j-1}$ is determined by the nominal class-size recurrence, so all
comparison-set sizes are fixed before sampling any instances.  To verify
that they fit in their suffix cubes, first bound the nominal sizes.  By
\eqref{eq:logsize} and \eqref{eq:achoice},
\[
  g_j\ \le\ g_{j-1}+(B-1)\Bigl(\frac{4g_{j-1}}B+1\Bigr)+1
  \ \le\ 5g_{j-1}+B.
\]
Since $g_0=d$, induction gives $g_J\le5^Jd+B(5^J-1)/4$, and since $B=o(d)$,
\begin{equation}\label{eq:g-small}
  d\ \le\ g_j\ \le\ K_Jd\qquad(0\le j\le J),\qquad K_J:=2\cdot5^J,
\end{equation}
for all sufficiently large $d$.  Thus $1\le a_j\le 4K_Jd/B+1\le Bt_{j-1}$,
since $B\to\infty$ and $t_{j-1}\ge d$.  The final nominal class size
satisfies $\log n_J=g_J=\Theta_\beta(d)$.

\subsection{Bounding the errors}

By \eqref{eq:alpha-j}, together with $a_j\ge4g_{j-1}/B$, $Bt_{j-1}\le\tau$ and
$g_{j-1}\ge d$,
\begin{equation}\label{eq:alpha-small}
  \alpha_j\ \le\ \sqrt2\,(\tau+1)\,2^{g_{j-1}/(B+1)-2g_{j-1}/B}
  \ \le\ \sqrt2\,(\tau+1)\,2^{-d/B}.
\end{equation}
By \eqref{eq:horizon} and $q=B$,
\begin{equation}\label{eq:t-asymp}
  \tau=d\bigl[1+B(B-1)\bigr]^J\sim dB^{2J}\sim d^{1+2J\theta}=d^{1/\beta}.
\end{equation}
Because $\theta<1$, we have $\log\tau=o(d/B)$.  Hence for every fixed
$C>0$ and every $Q\le\tau^C$,
\begin{equation}\label{eq:leak-vanish}
  \tau^2Q\max_j\alpha_j\ \le\ \sqrt2\,\tau^{C+2}(\tau+1)\,2^{-d/B}=o(1).
\end{equation}

For the overlap probability, use $a_j\le4K_Jd/B+1$, $g_{j-1}\le K_Jd$ and
$t_{j-1}\ge d$,
\[
  a_j+g_{j-1}-Bt_{j-1}\ \le\ K_Jd+\frac{4K_Jd}{B}+1-Bd=-\Omega(Bd),
\]
so \eqref{eq:collision-all} gives
\begin{equation}\label{eq:collision-vanish}
  \zeta\ \le\ \tau\,2^{-\Omega(Bd)}=o(1/\tau).
\end{equation}
Combining \eqref{eq:master}, \eqref{eq:leak-vanish} and
\eqref{eq:conditioning}, every learner making at most $\tau^C$ queries
incurs, under the distribution conditioned on no overlaps,
\begin{equation}\label{eq:many-mistakes}
  \EE[M_\tau\mid\mathcal E]\ \ge\ \frac{dB^J}{2}-o(1).
\end{equation}
Since $\tau\sim dB^{2J}$ and $\log n_J=\Theta_\beta(d)$, this is
$\Omega_\beta(\sqrt{\tau\log n_J})$.

\subsection{Prescribing the horizon and the class size exactly}

We now pad the conditioned construction to obtain each sufficiently large
$T$ and exactly $N=2^{\lfloor T^\beta\rfloor}$ traces as in \Cref{thm:main}.
Set $c_0:=1/(2K_J)=1/(4\cdot5^J)$ and run the construction with
\[
  d=\lfloor c_0T^\beta\rfloor.
\]
By \eqref{eq:g-small} and \eqref{eq:t-asymp}, $\tau\sim c_0^{1/\beta}T$ and
$g_J\le K_Jc_0T^\beta=T^\beta/2$; hence, for all sufficiently large $T$,
\begin{equation}\label{eq:padding-room}
  \tau+1\le T,\qquad g_J<\lfloor T^\beta\rfloor,\qquad \tau=\Theta_\beta(T).
\end{equation}
For each trace $h$ in the unpadded class, include the padded trace
$(0,h,0^{T-\tau-1})$, where $0^k$ denotes a string of $k$ zeros.  Include
also the lexicographically first $N-n_J$ traces of the form $(1,z)$ with
$z\in\cube^{T-1}$.  There are enough such traces since $N\le2^{T-1}$ for
large $T$.  The two sets are disjoint by their first bit, so the padded
class has exactly $N$ traces.  Its target is $(0,h^\star,0^{T-\tau-1})$,
where $h^\star$ is the unpadded target.

It remains to show that padding does not make the prediction problem easier. Given a learner for the padded class, we construct a learner for the original class that simulates its interaction. For each padded ERM query $w\in\RR^T$, the simulator queries the original oracle with $(w_2,\ldots,w_{\tau+1})$ to find the best trace beginning with $0$. It computes the best trace beginning with $1$ directly, since the added hypotheses are fixed and known. Comparing their scores and applying lexicographic tie-breaking reproduces the padded oracle's answer exactly. Thus, each padded query requires at most one query to the original oracle.

The simulator also supplies the known zero labels outside the original $\tau$ coordinates and uses the observed labels of $h^\star$ for the remaining rounds. Its predictions on these $\tau$ coordinates coincide with those of the padded learner. Hence, its mistake count equals the padded learner's mistake count on the original coordinates, and the padding preserves the expected mistake lower bound without increasing the query budget.

\begin{proof}[Proof of \Cref{thm:main}]
A padded learner making at most $T^C$ queries induces an unpadded learner
with the same cap.  By \eqref{eq:padding-room}, $T^C\le\tau^{C+1}$ for all
sufficiently large $T$, so \eqref{eq:many-mistakes} applies with exponent
$C+1$.  Since
$B=(1+o(1))\,d^\theta$ and $d=(1+o(1))\,c_0T^\beta$,
\[
  dB^J=(1+o(1))\,d^{(1+\beta)/(2\beta)}
  =(1+o(1))\,c_0^{(1+\beta)/(2\beta)}\,T^{(1+\beta)/2},
\]
while $\sqrt{T\log N}=\sqrt{T\lfloor T^\beta\rfloor}\le T^{(1+\beta)/2}$.
Thus under the padded distribution,
\[
  \EE[M_T]\ \ge\ c_\beta\sqrt{T\log N},\qquad
  c_\beta:=\tfrac14\,c_0^{(1+\beta)/(2\beta)},
\]
for all sufficiently large $T$.  The constant $c_\beta$ is independent of $C$.
For each learner, averaging gives a fixed pair $(\HH,h^\star)$ with
exactly $N$ traces on which the same bound holds in expectation over
the learner's randomness.
\end{proof}

\begin{remark}[Larger query budgets]\label{rem:budget}
The composition error in \eqref{eq:leak-vanish} is the only term in the
lower bound that depends on the query budget.  Since $\log\tau=o(d/B)$,
the error is still $o(1)$ when
$Q\le 2^{d/(2B)}$.  After padding, $d/B=\Theta_{\beta,J}
(T^{\beta-(1-\beta)/(2J)})$, and each padded query still costs one query.
Given $\varepsilon>0$, choose $J$ with
$(1-\beta)/(2J)<\min\{\beta,\varepsilon\}$.  The same construction then
gives the lower bound of \Cref{thm:main} for budgets of up to
$2^{T^{\beta-\varepsilon}}$ queries, with a positive constant depending
on $\beta$ and $\varepsilon$.  At a budget of $2TN$ queries, enumeration
followed by halving achieves $\log N$ mistakes (\Cref{rem:enumerate}).
\end{remark}

\section{Separating halving from weighted ERM for small classes}\label{sec:small}

We now show that the separation persists even when the hypothesis class is much smaller, with $\log N=\Theta((\log T)^2)$. Unlike \Cref{thm:main}, the following result is established along a sequence of horizons and class sizes.

\begin{theorem}[Separation for small classes]\label{thm:small}
There are sequences of horizon and class-size pairs $(T_i,N_i)_{i\ge1}$
with $T_i\to\infty$ of both kinds described below. Fix such a sequence and
$C>0$. For all sufficiently large $i$, every randomized learner $\mathcal A$
making at most $T_i^C$ weighted-ERM queries on every execution has a hard
instance satisfying the stated lower bound. Here a hard instance consists
of a class $\HH\subseteq\cube^{T_i}$ of exactly $N_i$ distinct traces and
a target $h^\star\in\HH$.
\begin{enumerate}[label=(\alph*),leftmargin=20pt,topsep=2pt,itemsep=2pt]
\item For every $\eta\in(0,1/2)$, the sequence has
  $\log N_i=O((\log T_i)^p)$ for some $p=p(\eta)<\infty$, and
  \[
    \EE_{\mathcal A}[M_{T_i}]\ \ge\ T_i^{1/2-\eta}.
  \]
\item The sequence has $\log N_i=\Theta((\log T_i)^2)$, and for some
  absolute constant $c>0$,
  \begin{equation}\label{eq:smallclaim}
    \EE_{\mathcal A}[M_{T_i}]\ \ge\ 2^{c\log T_i/\log\log T_i}.
  \end{equation}
\end{enumerate}
\end{theorem}

\begin{proof}
We let the recursion depth $J$ grow with the horizon. The factor $K_J$ in
the fixed-depth estimate $g_J\le K_Jd$ grows exponentially in $J$, so we
bound the nominal log class size $g_J$ directly from \eqref{eq:logsize}.

\paragraph{Part (a).}
Fix a constant $\kappa>0$ and let $J\to\infty$.  Choose
\begin{equation}\label{eq:small-params}
  q=\lceil J^\kappa\rceil,\quad
  B=\text{the least odd integer }\ge32Jq,\quad
  d=\bigl\lceil J\bigl(\log(2Jq)\bigr)^2\bigr\rceil,\quad
  a_j=d\ \text{for all }j.
\end{equation}
Since $t_{j-1}\ge d$, the comparison-set exponent $a_j=d$ satisfies
$1\le a_j\le Bt_{j-1}$. By \eqref{eq:logsize},
$g_{j-1}\le d+(j-1)((q-1)d+1)\le d+Jq(d+1)$. Because $B\ge32Jq$, this implies
$g_{j-1}/(B+1)\le d/16$ for large $J$. Substituting into \eqref{eq:alpha-j}
and using $Bt_{j-1}\le\tau$ gives
\begin{equation}\label{eq:small-alpha}
  \alpha_j\ \le\ \sqrt2\,(\tau+1)\,2^{-7d/16}.
\end{equation}
The horizon formula \eqref{eq:horizon} and the class-size bounds
\eqref{eq:logsize} give
\begin{align}
  \log\tau&=\log d+J\log\bigl(1+(q-1)B\bigr)=(2\kappa+1+o(1))\,J\log J,
  \label{eq:small-logt}\\
  g_J&=\Theta(Jqd)=\Theta\bigl(J^{\kappa+2}(\log J)^2\bigr).\label{eq:small-g}
\end{align}
In particular, $\log\tau=o(d)$ and $g_J=O((\log\tau)^{\kappa+2})$.
By \eqref{eq:small-alpha}, the composition error in \eqref{eq:master} is
$\tau^2Q\max_j\alpha_j=o(1)$ for every budget $Q\le\tau^C$ with $C$ fixed.

To obtain exactly $n_J$ traces, we condition on the event $\mathcal E$ of
no overlaps from \Cref{subsec:collisions}. Since
$a_j+g_{j-1}-Bt_{j-1}\le d+(B+1)d/16-Bd=-\Omega(Bd)$,
\eqref{eq:collision-all} gives $\zeta\le\tau 2^{-\Omega(Bd)}=o(1/\tau)$.
By \eqref{eq:master} and \eqref{eq:conditioning}, every learner with budget
$Q\le\tau^C$ thus satisfies
\[
  \EE[M_\tau\mid\mathcal E]\ \ge\ \frac{dq^J}{2}-o(1).
\]
The growth exponent of $dq^J$ in the horizon satisfies
\begin{equation}\label{eq:small-exponent}
  \frac{\log(dq^J)}{\log\tau}\ \longrightarrow\ \frac{\kappa}{2\kappa+1}.
\end{equation}
Given $\eta\in(0,1/2)$, choose $\kappa$ large enough that
$\kappa/(2\kappa+1)>1/2-\eta$. The lower bound then exceeds
$\tau^{1/2-\eta}$ for all sufficiently large $J$, while
$\log n_J=O((\log\tau)^p)$ with $p=\kappa+2$. Taking $(\tau,n_J)$ as the
sequence of horizon and size pairs and averaging over the conditioned
distribution yields a fixed hard instance for each learner.

\paragraph{Part (b).}
Now take $q=2$ and keep the definitions of $B$, $d$ and $a_j$ in
\eqref{eq:small-params}. The same estimates for $\alpha_j$ and $\zeta$ apply,
and the horizon and log class size now satisfy
\[
  \log\tau=\Theta(J\log J),\qquad
  g_J=\Theta(Jd)=\Theta\bigl(J^2(\log J)^2\bigr)=\Theta\bigl((\log\tau)^2\bigr).
\]
For every fixed $C$ the conditional mistake lower bound is $d2^{J-1}-o(1)$.
Since $J=\Theta(\log\tau/\log\log\tau)$, this is at least
$2^{c\log\tau/\log\log\tau}$ for some absolute constant $c>0$ and all
sufficiently large $J$. Averaging on $\mathcal E$ gives a fixed hard
instance with horizon $\tau$ and exactly $n_J$ traces for each learner.
\end{proof}

\section{Related work}\label{sec:related}

\paragraph{Transductive online prediction.}
The transductive online learning model, in which the instance sequence is known in advance, was studied by \citet{ben-david-kushilevitz-mansour}; \citet{kakade-kalai} consider a related variant in which only the instance set, rather than its order, is known. For a finite realizable class of size $N$, the classical halving algorithm makes at most $\log_2 N$ mistakes when the consistent hypotheses can be counted \citep{barzdin1972prediction,angluin1988queries,littlestone}. Random playout and randomized rounding achieve the agnostic-regret rate $O(\sqrt{T\log N})$ using $O(T)$ optimization calls \citep{cesabianchi-shamir-nips11,cesabianchi-shamir,rakhlin-shamir-sridharan}. Related work characterizes transductive mistake bounds in terms of VC and Littlestone dimensions for explicitly represented hypothesis classes \citep{hanneke-moran-shafer,chase-hanneke-moran-shafer}. In contrast, our results establish the limitations of optimization-oracle access: even polynomially many weighted-ERM queries cannot recover the logarithmic mistake bound of halving.

\paragraph{Oracle complexity of realizable learning.}
Consistency and ERM oracles for online learning have been studied by
\citet{assos-et-al,kozachinskiy-steifer,daskalakis-golowich}.
The closest results in the transductive setting are due to
\citet{attias-hanneke-ramaswami}.  They study ERM queries on
labeled examples and a weaker consistency oracle that reports whether
a labeled sample is realizable.  Their lower bounds apply to
deterministic learners making fewer than $T/2$ ERM queries, and to
randomized learners making at most $T/20$ weak-consistency queries.
In contrast, \Cref{thm:main} covers randomized learners making
\emph{any fixed polynomial} number of queries to a stronger weighted-ERM
oracle, which accepts arbitrary signed real objectives and returns
an entire maximizing hypothesis.

\citet{attias-hanneke-ramaswami} also ask whether randomized learners
can achieve $O(\dld)$ or $O(\dvc\log T)$ mistakes using
$T^C 2^{O(\dvc)}$ queries, where $\dld$ and $\dvc$ denote the
Littlestone and VC dimensions and $C$ is independent of the class.
Our lower bound establishes that a dimension-dependent enlargement
of a polynomial query budget is necessary for such guarantees in
general.  Indeed, the hard classes in \Cref{thm:main} have
$\dvc=\Theta_\beta(T^\beta)$: their projection onto the coordinates
of an initial $d$-dimensional full cube is the whole cube,
with $d=\Theta_\beta(T^\beta)$, and $\dvc\le\log_2N$.
Therefore their $\Omega_\beta(T^{(1+\beta)/2})$ expected mistake lower
bound exceeds $O(\dvc\log T)$ for every fixed $\beta<1$.
The budget $T^C2^{O(\dvc)}$ can, however, be exponentially larger
than a polynomial in $T$ on these classes, so the dimension-dependent
question concerns a different query-budget regime.
Oracle selection rules have also been studied by \citet{li-2026},
while \citet{attias-hanneke-ramaswami-2026} consider
agnostic, non-transductive learning.

\paragraph{Online algorithms from optimization.}
Follow-the-Perturbed-Leader (FTPL) and related methods use offline optimization oracles to construct online learning algorithms \citep{kalai-vempala,syrgkanis-krishnamurthy-schapire,dudik-et-al}. Lower bounds on the power of optimization-oracle access in related settings have been established by \citet{hazan-koren} and \citet{block-dagan-golowich-rakhlin}. On the positive side, small-loss oracle-efficient algorithms \citep{wang-hu-muthukumar-abernethy,block-shetty} demonstrate that additional structural assumptions can lead to stronger guarantees in realizable settings. For arbitrary finite trace classes, however, the first-order bounds of \citet{wang-hu-muthukumar-abernethy} still contain an additive term of agnostic order. Our lower bound shows that this dependence is unavoidable in general, even with polynomially many weighted-ERM queries.

\paragraph{Restricted instance generation.}
Under i.i.d., hybrid, or smoothed assumptions, the instance sequences possess additional structure that optimization-based algorithms can exploit \citep{lazaric-munos,wu-sima-szpankowski,okoroafor-kleinberg-kim,buzaglo-hazan,haghtalab-han-shetty-yang,block-rakhlin-shetty}. In contrast, we study worst-case transductive online learning, where the instance sequence is known in advance but no distributional or smoothness assumptions are imposed. The main analytic tool in our proof is Boolean hypercontractivity \citep{bonami1970etude,beckner1975inequalities,odonnell}, for which we provide a self-contained proof of the required inequality in \Cref{app:hyper}.

\section{Discussion and open problems}\label{sec:discussion}

For $N=2^{\lfloor T^\beta\rfloor}$ with any fixed $\beta\in(0,1)$, our results establish that the worst-case realizable mistake bound achievable with polynomially many weighted-ERM queries is $\Theta(\sqrt{T\log N})$, matching the agnostic rate despite the $\log_2 N$ mistake bound of halving with explicit class access. The separation persists for much smaller classes, although the optimal polynomial-query mistake bound in that regime remains open. The key mechanism is the composition lemma (Lemma~\ref{lem:compose}), which hides each new prediction problem behind a fixed comparison set, preserving its additive contribution to the mistake lower bound while introducing its hypotheses at only one realized prefix.


\begin{remark}[Expected query budgets]\label{rem:expected}
Our lower bounds also apply, up to an additive constant, to learners whose
\emph{expected} number of queries is at most $Q$ on every instance.
Indeed, stopping such a learner after $TQ$ queries, and predicting $0$ on
all remaining rounds without further queries, yields a $TQ$-query learner;
by Markov's inequality, this changes the expected number of mistakes by at
most $T\cdot\PP(\text{more than }TQ\text{ queries})\le1$.
\end{remark}

\begin{remark}[Some restriction is necessary]\label{rem:enumerate}
A weighted-ERM query can determine whether a given prefix $z\in\cube^k$ extends to a hypothesis in $\HH$. Indeed, set $w_i=2z_i-1$ for $i\le k$ and $w_i=0$ otherwise. Then
\[
  \langle w,h\rangle
  =\|z\|_1-|\{i\le k:h_i\ne z_i\}|,
\]
so the optimal value equals $\|z\|_1$ if and only if some $h\in\HH$ agrees with $z$ on the first $k$ coordinates. By exploring the tree of extendable prefixes, as in \citet[Lemma~4.1]{attias-hanneke-ramaswami}, we can recover the entire class using at most $2TN$ queries: at each depth, there are at most $N$ extendable prefixes, each with two possible extensions. Once $\HH$ is recovered, halving achieves at most $\log_2N$ mistakes. Thus, when $N$ is polynomial in $T$, polynomially many weighted-ERM queries suffice to match halving. More generally, a restriction on the query budget is essential for any separation.
\end{remark}

\paragraph{Open questions.}
For classes with $\log N=\Theta((\log T)^2)$, our lower bound is $\exp(\Omega(\log T/\log\log T))$, while random playout achieves $O(\sqrt{T}\log T)$. Closing this gap remains an open problem. Another question is to characterize the tradeoff between the number of oracle queries and the achievable mistake bound, bridging the lower bound in \Cref{rem:budget} and the enumeration upper bound above. Finally, it would be interesting to identify structural assumptions on the hypothesis class or additional oracle capabilities under which logarithmic mistake bounds become achievable with polynomially many queries.

\newpage
\section*{Statement on the Use of AI Tools}

AI systems made substantial contributions to this research, which we disclose here for transparency. Initial investigations with OpenAI's GPT-5.6 Sol and Anthropic's Fable 5 did not resolve the problem. OpenAI's GPT-6 Astra subsequently established a weaker lower bound of order $\Omega(\log N\cdot\log T)$. The main lower bound and its proof presented in this paper were developed by GPT-6.2 Astra. The author formulated the research problem, guided the investigation, and independently verified the mathematical arguments. Claude Code and Codex assisted with manuscript preparation and editing. The author assumes full responsibility for the correctness and content of the paper.

\appendix
\section{An elementary proof of the hypercontractive inequality}\label{app:hyper}

For completeness, we prove the only form of hypercontractivity used in
\Cref{lem:hiding}, following the standard route of a one-bit inequality,
tensorization and duality, as in \citet{bonami1970etude}
\citep[cf.][Sections~9.3--9.4]{odonnell}.  Identify $\cube$ with
$\{-1,+1\}$ via $x\mapsto(-1)^x$; this preserves the uniform measure, the
Fourier coefficients and all $L_p$ norms, and turns the characters $\chi_U$
into monomials.  Let $B=2\ell-1\ge3$ be odd and $\rho=B^{-1/2}$, and let
$\Tn_\rho$ multiply the Fourier coefficient indexed by $U$ by $\rho^{|U|}$.
We prove that, for every $f:\{-1,+1\}^r\to\RR$,
\begin{equation}\label{eq:app-hyper}
  \|\Tn_\rho f\|_2\ \le\ \|f\|_{1+1/B}.
\end{equation}

\paragraph{One bit.}
For real $a,b$, a uniformly random sign $\sigma$, and $\ell\ge2$,
\begin{equation}\label{eq:onebit}
  \EE_\sigma\Bigl|a+\frac{\sigma b}{\sqrt{2\ell-1}}\Bigr|^{2\ell}
  \ \le\ (a^2+b^2)^\ell.
\end{equation}
Indeed, the odd powers of $\sigma$ average to zero, so the left-hand side
equals
\[
  \sum_{i=0}^{\ell}\binom{2\ell}{2i}\frac{a^{2\ell-2i}b^{2i}}{(2\ell-1)^i}.
\]
For every $0\le i\le\ell$,
\begin{equation}\label{eq:bin-coef}
  \binom{2\ell}{2i}\ \le\ \binom\ell i\,(2\ell-1)^i.
\end{equation}
To see this, note that the ratio of the left-hand side to $\binom\ell i$
equals, with the convention $(-1)!!=1$,
\[
  \frac{(2\ell-1)!!}{(2i-1)!!\,(2\ell-2i-1)!!}
  =\frac{\prod_{j=0}^{i-1}(2\ell-2j-1)}{(2i-1)!!}
  \ \le\ (2\ell-1)^i.
\]
Substituting \eqref{eq:bin-coef} and applying the binomial theorem yields
\eqref{eq:onebit}.

\paragraph{Tensorization.}
We prove by induction on $r$ the dual estimate
$\|\Tn_\rho u\|_{2\ell}\le\|u\|_2$.  The case $r=0$ is trivial.  Write a
function of $r$ signs as $u(z,\sigma)=a(z)+\sigma b(z)$, where $z$ denotes
the first $r-1$ signs, and let $\tilde a=\Tn_\rho a$ and $\tilde b=\Tn_\rho b$,
with the noise operator acting on the first $r-1$ coordinates.  Then
$\Tn_\rho u(z,\sigma)=\tilde a(z)+\rho\sigma\tilde b(z)$.  Applying
\eqref{eq:onebit} pointwise, then Minkowski's inequality in $L_\ell$, and
then the induction hypothesis,
\begin{align*}
  \|\Tn_\rho u\|_{2\ell}^2
  &=\Bigl(\EE_z\EE_\sigma\bigl|\tilde a(z)+\rho\sigma\tilde b(z)\bigr|^{2\ell}\Bigr)^{1/\ell}
   \le\Bigl(\EE_z\bigl(\tilde a(z)^2+\tilde b(z)^2\bigr)^\ell\Bigr)^{1/\ell}\\
  &\le\|\tilde a\|_{2\ell}^2+\|\tilde b\|_{2\ell}^2
   \le\|a\|_2^2+\|b\|_2^2=\|u\|_2^2.
\end{align*}
This proves $\|\Tn_\rho u\|_{2\ell}\le\|u\|_2$ on every Boolean cube.

\paragraph{Duality.}
The operator $\Tn_\rho$ is self-adjoint with respect to the uniform measure,
and $p=1+1/B$ has conjugate exponent $p'=B+1=2\ell$.  For every $f$, by
H\"older's inequality and the tensorized bound,
\[
  \|\Tn_\rho f\|_2
  =\sup_{\|u\|_2\le1}\EE\bigl[(\Tn_\rho f)\,u\bigr]
  =\sup_{\|u\|_2\le1}\EE\bigl[f\,\Tn_\rho u\bigr]
  \le\|f\|_p\sup_{\|u\|_2\le1}\|\Tn_\rho u\|_{p'}
  \le\|f\|_p,
\]
which is \eqref{eq:app-hyper}.

\bibliographystyle{plainnat}
\bibliography{references}

\end{document}